\documentclass[letterpaper, 10 pt, conference]{ieeeconf}  

\IEEEoverridecommandlockouts                              

\usepackage{amsmath}       
\usepackage{amssymb}       
\usepackage{booktabs}      
\usepackage{multirow}
\usepackage{makecell}      
\usepackage{xcolor}        
\usepackage{caption}
\usepackage[normalem]{ulem}
\usepackage{array}

\usepackage{graphicx}
\usepackage{subcaption}
\usepackage{adjustbox}
\usepackage{algorithm}
\usepackage{algpseudocode}
\usepackage{paralist}

\newcommand{\best}[1]{\textbf{#1}}
\newcommand{\secondbest}[1]{\underline{#1}}

\newcommand{\vci}[2][]{\raisebox{-0.5\height}{\includegraphics[#1]{#2}}}
\usepackage{hyperref}
\usepackage{colortbl}
\usepackage{xargs}
\usepackage{xcolor}  
\usepackage[colorinlistoftodos,prependcaption,textsize=tiny]{todonotes}
\newcommandx{\Shida}[2][1=]{\todo[inline,linecolor=blue,backgroundcolor=blue!25,bordercolor=blue,#1]{#2}}
\newcommandx{\Sen}[2][1=]{\todo[inline,linecolor=red,backgroundcolor=red!25,bordercolor=red,#1]{#2}}

\title{\LARGE \bf
NAS-GS: Noise-Aware Sonar Gaussian Splatting 
}

\author{Shida Xu, Jingqi Jiang, Jonatan Scharff Willners, and Sen Wang}
\author{Shida Xu$^{1}$, Jingqi Jiang$^{1}$, Jonatan Scharff Willners$^{2}$, and Sen Wang$^{1}$
\thanks{$^{1}$ I-X and Department of Electrical and Electronic Engineering, Imperial College London, UK
{\tt\small \{s.xu23, j.jiang23, sen.wang\}@imperial.ac.uk}}%
\thanks{$^{2}$Frontier Robotics, The National Robotarium, Edinburgh UK
}%
}

\UseRawInputEncoding
\begin{document}

\maketitle
\thispagestyle{empty}
\pagestyle{empty}

\begin{abstract}

Underwater sonar imaging plays a crucial role in various applications, including autonomous navigation in murky water, marine archaeology, and environmental monitoring. However, the unique characteristics of sonar images, such as complex noise patterns and the lack of elevation information, pose significant challenges for 3D reconstruction and novel view synthesis. In this paper, we present NAS-GS, a novel Noise-Aware Sonar Gaussian Splatting framework specifically designed to address these challenges. Our approach introduces a Two-Ways Splatting technique that accurately models the dual directions for intensity accumulation and transmittance calculation inherent in sonar imaging, significantly improving rendering speed without sacrificing quality. Moreover, we propose a Gaussian Mixture Model (GMM) based noise model that captures complex sonar noise patterns, including side-lobes, speckle, and multi-path noise. This model enhances the realism of synthesized images while preventing 3D Gaussian overfitting to noise, thereby improving reconstruction accuracy. We demonstrate state-of-the-art performance on both simulated and real-world large-scale offshore sonar scenarios, achieving superior results in novel view synthesis and 3D reconstruction.

\end{abstract}

\section{Introduction}
Underwater inspections are vital for maintaining offshore infrastructure, conducting marine archaeology, and monitoring the environment. While optical cameras are effective in clear waters, imaging sonar systems offer robust perception in turbid environments where light penetration is limited. However, the distinct nature of acoustic imaging characterized by complex noise patterns, low resolution, and nonlinear geometry presents substantial hurdles for 3D reconstruction and novel view synthesis.

Recent progress in differentiable rendering, such as Neural Radiance Fields (NeRF) and Gaussian Splatting (GS) \cite{mildenhall2021nerf, barron2022mip, fridovich2022plenoxels, kerbl20233d, guedon2024sugar}, has enabled photorealistic view synthesis and geometric reconstruction for optical cameras. Yet, directly applying these methods to sonar imaging is ineffective due to fundamental differences in sensing modalities. Sonar images possess unique properties, including intensity accumulation along elevation arcs, range-dependent transmittance, and significant noise artifacts like side-lobes, speckle, and multi-path reflections. These factors demand specialized geometric modeling and robust noise handling strategies.

Differentiable rendering for sonar has recently gained traction. Early works utilized NeRF and signed distance functions (SDF) for sonar synthesis and reconstruction. Neusis \cite{qadri2023neural, neusis_ngp} proposed a physics-based sonar rendering model within a NeRF framework, facilitating novel view synthesis and 3D reconstruction. Subsequent work \cite{lin2025acoustic} extended this to handle pose drift. However, NeRF-based approaches are often computationally intensive and slow, limiting real-time utility. Furthermore, they frequently struggle to model complex sonar noise, resulting in reduced reconstruction quality on real-world scenarios with noisy data. Differentiable Space Carving (DSC) \cite{dsc_sonar} offers a more efficient alternative but lacks the ability to model complex noise characteristics. As noted in \cite{sethuraman2025sonarsplat}, both NeRF-based methods and DSC suffer from limited quality in novel view synthesis.

\begin{figure}
\centering
\includegraphics[width=1\columnwidth]{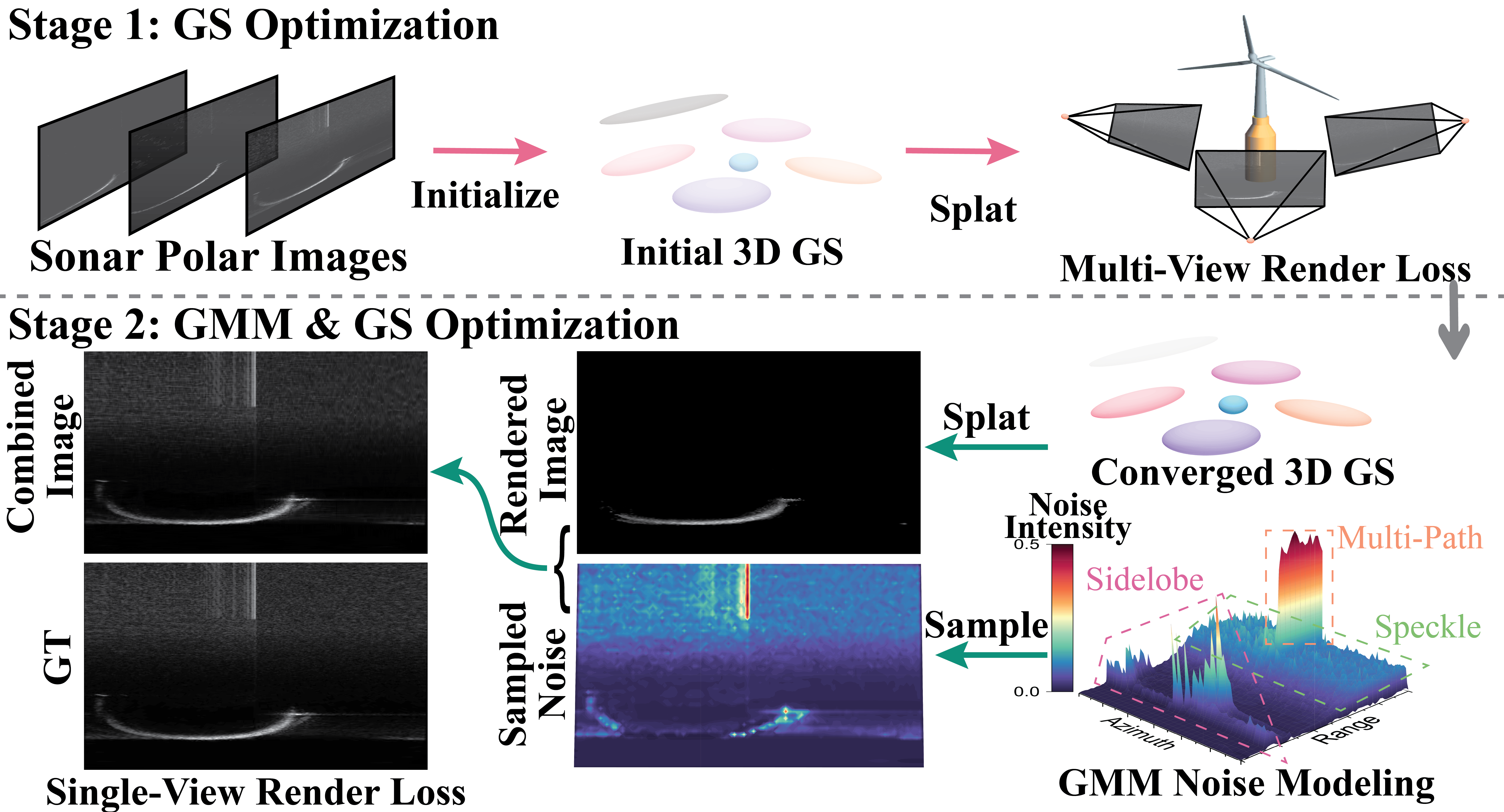}
\caption{Overview of the proposed NAS-GS pipeline.}
\label{fig:pipeline}
\end{figure}

Gaussian Splatting offers faster rendering speeds than NeRF due to its explicit 3D representation, making it suitable for real-time applications. Recent efforts have adapted GS to sonar and fuse with camera data. ZSplat \cite{qu2024z} uses a simplified sonar geometry model, which limits its efficacy in real-world scenarios. Aqua-Splat \cite{ling2025aqua} incorporates a more accurate physics-based forward model but lacks explicit noise modeling and relies on computationally expensive volume rendering. This approach, while adequate for small, controlled environments, may be inefficient for large-scale offshore scenes. Both methods also exhibit performance degradation without camera input due to insufficient noise modeling and have primarily been validated on simulated or small-scale tank data.

SonarSplat \cite{sethuraman2025sonarsplat} is the most relevant prior work, focusing on sonar-based Gaussian Splatting. Although it introduces a sonar-specific pipeline and a noise model for azimuth streaking, it has notable limitations: (1) its noise model simplifies noise as per-pixel gain, which handles dark azimuth streaking but fails to capture complex side-lobe, speckle, and multi-path noise; and (2) its rendering speed is considerably slower than camera-based methods. These issues hinder its application in large-scale offshore environments where complex noise is prevalent and efficiency is paramount.

To overcome these limitations, we propose NAS-GS, a novel Noise-Aware Sonar Gaussian Splatting framework as shown in Figure \ref{fig:pipeline}, designed for accurate sonar imaging through rigorous geometric modeling and comprehensive noise characterization. Our main contributions are:

\begin{itemize}
    \item A novel Two-Ways Splatting technique that accurately models the unique projection characteristics of imaging sonar. It efficiently manages the dual directions of intensity accumulation and transmittance calculation, boosting rendering speed to over 700 Frames Per Second (FPS) without compromising quality.
    \item A learnable Gaussian Mixture Model (GMM) based noise model that captures complex sonar noise patterns, including side-lobes, speckle, and multi-path noise. This model improves the realism of synthesized images and prevents 3D Gaussians from overfitting to noise, thereby enhancing reconstruction accuracy. Furthermore, our framework supports rendering both realistic noisy images and denoised images for simulation and data augmentation.
    \item We demonstrate state-of-the-art performance on both simulated and real-world large-scale offshore sonar datasets, achieving superior results in novel view synthesis and 3D reconstruction.
\end{itemize}

We will release our code, simulation dataset upon the paper's acceptance. For more details, please visit our project page\footnote{\url{https://senseroboticslab.github.io/NAS-GS-Page/}}.

\section{The Proposed NAS-GS Method}

\subsection{Sonar Geometry}\label{sec:sonar_geometry}

\begin{figure}[htbp]
    \centering
    \begin{subfigure}[b]{\linewidth}
        \centering
        \includegraphics[width=0.99\linewidth]{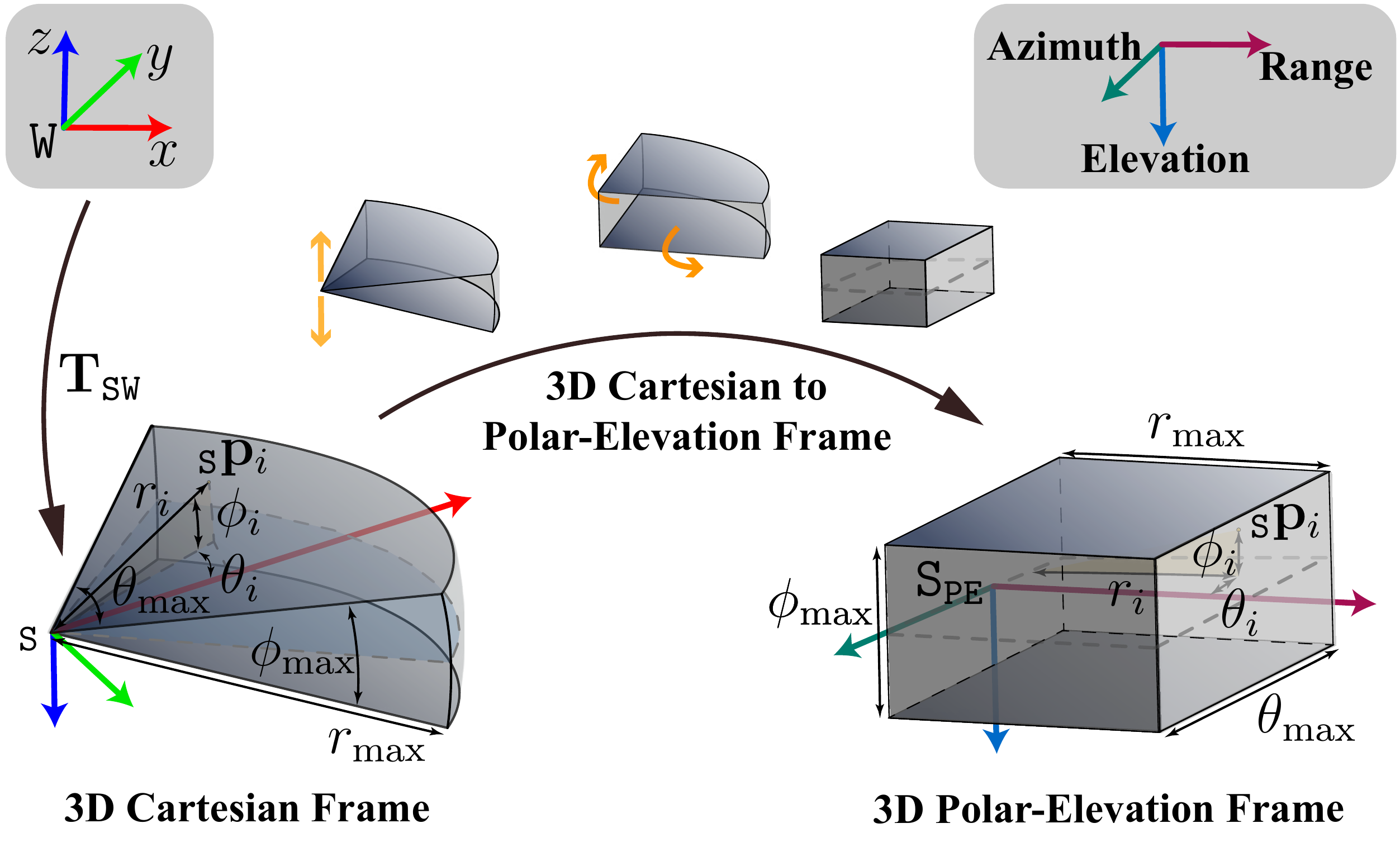}
        \caption{Sonar coordinate frames and transformations}
        \label{fig:sonar_model}
    \end{subfigure}
    
    \vspace{0.3cm}
    
    \begin{subfigure}[b]{\linewidth}
        \centering
        \includegraphics[width=0.99\linewidth]{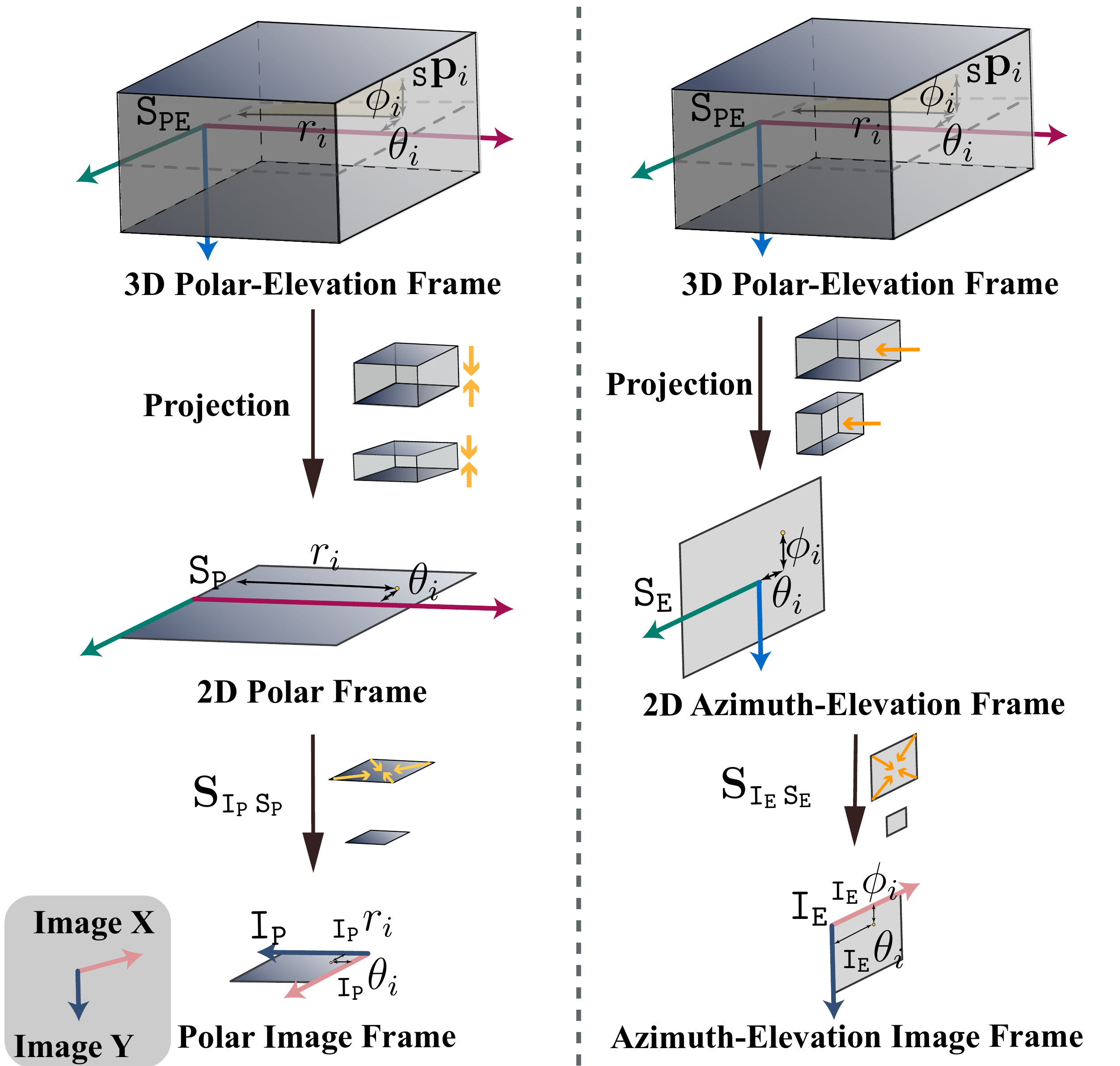}
        \caption{Sonar projection from 3D to 2D image space}
        \label{fig:sonar_projection}
    \end{subfigure}
    \caption{Sonar Geometry Model}
    \label{fig:sonar_geometry}
\end{figure}

To formulate sonar geometry, we define multiple coordinate frames: the world frame ($\mathtt{W}$), the sonar Cartesian frame ($\mathtt{S}$), an intermediate 3D representation sonar polar-elevation ($\mathtt{S_{PE}}$), and two intermediate 2D representations sonar polar ($\mathtt{S_P}$) and sonar elevation-azimuth ($\mathtt{S_E}$) which are subsequently mapped to their respective image pixel coordinates ($\mathtt{I_P}$ and $\mathtt{I_E}$). 
The transformation pipeline consists of a rigid body transformation from world to sonar Cartesian coordinates, followed by a non-linear transformation to polar-elevation coordinates, then projection functions that convert 3D polar-elevation coordinates into 2D representations, and finally similarity transformations that map these representations to pixel coordinates.
The Sonar Geometry model is shown in Figure \ref{fig:sonar_geometry}.

\subsubsection{Coordinate Frame Definitions}

A point in the world frame and the sonar frame are denoted as:
\begin{equation}
 _\mathtt{W}\mathbf{p}_{i} \doteq \begin{bmatrix}x \\ y \\ z\end{bmatrix} \quad \text{and} \quad
  _\mathtt{S}\mathbf{p}_{i} \doteq \begin{bmatrix}x_s \\ y_s \\ z_s\end{bmatrix}
\end{equation}


\subsubsection{World to Sonar Transformation}

The transformation from the world frame to the sonar frame is given by the homogeneous transformation matrix:
\begin{equation}
\mathbf{T}_{\mathtt{S}\mathtt{W}} \doteq \begin{bmatrix}\mathbf{R}_{\mathtt{S}\mathtt{W}}& _{\mathtt{S}}\mathbf{t}_{\mathtt{S}\mathtt{W}} \\ \mathbf{0} &1\end{bmatrix}
\end{equation}
where $\mathbf{R}_{\mathtt{S}\mathtt{W}}$ is the rotation matrix and $_{\mathtt{S}}\mathbf{t}_{\mathtt{S}\mathtt{W}}$ is the translation vector. A point in the sonar frame is obtained by:
\begin{equation}
 _\mathtt{S}\mathbf{p}_{i} =\mathbf{T}_{\mathtt{S}\mathtt{W}} \, _{\mathtt{W}}\mathbf{p}_{i}
\end{equation}

\subsubsection{Cartesian to Polar-Elevation Conversion}
A 3D point in the sonar frame can be projected into a 3D polar-elevation coordinate system, whose x and y stand for the azimuth angle and range respectively, and z stands for the elevation angle. The projection process is first to shift the origin along the z axis, then rotate the elevation-range planes of both side as shown in figure \ref{fig:sonar_projection}. Therefore, the function $\pi_{pe}$ that maps a 3D Cartesian point in the sonar frame to polar-elevation coordinates is
\begin{equation}
  \begin{aligned}
    _{\mathtt{S_{PE}}}\mathbf{p}_{i} &= \pi_{pe}(\, _{\mathtt{S}}\mathbf{p}_{i}) \\
    &= \begin{bmatrix} 
    r_{i}\\
    \theta_{i}\\
    \phi_{i}
    \end{bmatrix}
    = \begin{bmatrix}
    \sqrt{x_s^{2}+y_s^{2}+z_s^{2}}\\
    \text{atan2}(y_s, x_s) \\
    \arcsin\left(\frac{z_s}{\sqrt{x_s^{2}+y_s^{2}+z_s^{2}}}\right) 
  \end{bmatrix}
  \end{aligned}
\end{equation}
where ${r_{i}}$ represents the range, ${\theta_{i}}$ represents the azimuth angle, and ${\phi_{i}}$ represents the elevation angle.

\subsubsection{3D Polar-Elevation to 2D Polar Image Projection}

The projection function $\pi_{p}$ maps a 3D polar-elevation point to a 2D polar representation by discarding the elevation component:
\begin{align}
\pi_{p}( _{\mathtt{S_{PE}}}\mathbf{p}_{i}) \doteq 
{_{\mathtt{S_P}}\hat{\mathbf{p}}_{i}}
=
\begin{bmatrix} 
r_{i}\\
\theta_{i}
\end{bmatrix}
\end{align}


\paragraph{Polar to Image Pixel Transformation}

The transformation from the polar representation to image pixel coordinates applies a similarity transformation that scales the coordinates, translates the polar frame origin to the image frame origin (located at the top-left corner), and rotates the frame to align the image x-axis with azimuth and the y-axis with range. This is defined as:
\begin{align}
 _{\mathtt{I_P}}\mathbf{p}_{i} 
&= 
\mathbf{S}_{\mathtt{I_P}\,\mathtt{S_P}} \, _{\mathtt{S_P}}\hat{\mathbf{p}}_{i} 
= 
\begin{bmatrix}
{}_{\mathtt{I_P}}r_{i}\\
{}_{\mathtt{I_P}}\theta_{i}
\end{bmatrix}\nonumber\\
&= \begin{bmatrix} 
  s_{\mathtt{r}} \cos(\omega) & -s_{\theta} \sin(\omega)&t_\mathrm{x} \\ 
  s_{\mathtt{r}} \sin(\omega) & s_{\theta} \cos(\omega)&t_\mathrm{y}\\
0&0&1
\end{bmatrix}
\begin{bmatrix} 
r_{i}\\
\theta_{i}\\
1
\end{bmatrix}
\end{align}
where $s_{\mathtt{r}}$ and $s_{\theta}$ are scale factors, $\omega$ is a angle offset, and $t_\mathrm{x}$, $t_\mathrm{y}$ are translation offsets. 

\subsubsection{3D Polar-Elevation to 2D Elevation-Azimuth Image Projection}

The projection function $\pi_{e}$ maps a 3D polar-elevation point to a 2D elevation-azimuth representation by discarding the range component:
\begin{align}
\pi_{e}( _{\mathtt{S_{PE}}}\mathbf{p}_{i}) &\doteq {_{\mathtt{S_E}}\hat{\mathbf{p}}_{i}} = \begin{bmatrix}  \phi_{i} \\ \theta_{i}  \end{bmatrix}
\end{align}


\paragraph{Elevation-Azimuth to Image Pixel Transformation}

The transformation from the elevation-azimuth representation to image pixel coordinates applies a similarity transformation similar to the polar case. This is defined as:
\begin{align}
 _{\mathtt{I_E}}\mathbf{p}_{i} 
&= 
\mathbf{S}_{\mathtt{I_E}\,\mathtt{S_E}} \, _{\mathtt{S_E}}\hat{\mathbf{p}}_{i} 
= 
\begin{bmatrix}
  {}_{\mathtt{I_E}}\phi_{i}\\
{}_{\mathtt{I_E}}\theta_{i}
\end{bmatrix}\\
&=
\begin{bmatrix} 
s_{\phi} \cos(\omega)& -s_{\theta} \sin(\omega)&t_\mathrm{x}\\ 
s_{\phi}\sin(\omega)& s_{\theta}\cos(\omega)&t_\mathrm{y}\\
0&0&1
\end{bmatrix}
\begin{bmatrix} 
\phi_{i} \\  
\theta_{i} \\ 
1 
\end{bmatrix}
\end{align}
where $s_{\phi}$ and $s_{\theta}$ are scale factors, $\omega$ is a rotation angle, and $t_\mathrm{x}$, $t_\mathrm{y}$ are translation parameters.

\subsection{Sonar Gaussian Splatting}

\subsubsection{Gaussian Representation}
We revised the Gaussian representation in \cite{kerbl20233d}. Each Gaussian is defined under world frame ($\mathtt{W}$) as follows:
\begin{equation}
\mathcal{G}_i = \{\boldsymbol{\mu}_i, \boldsymbol{\Sigma}_i, I_i, \Lambda_i\}
\end{equation}
where $\boldsymbol{\mu}_i \doteq [x,y,z]^\mathtt{T} \in \mathbb{R}^3$ is the mean position, $\boldsymbol{\Sigma}_i \in \mathbb{R}^{3\times3}$ is the covariance matrix, $I_i \in \mathbb{R}^3$ is the color intensity, and $\Lambda_i \in \mathbb{R}$ is the opacity of the $i$-th Gaussian. 

\subsubsection{Mean Projection}

The mean position of each Gaussian is projected into the sonar image space following the Sonar Geometry in \ref{sec:sonar_geometry}. 

\paragraph{Polar Image}
For polar images, the projected mean in pixel coordinates is obtained by the complete transformation chain:
\begin{equation}
 _\mathtt{I_P}\boldsymbol{\mu}_{i} = \mathbf{S}_{\mathtt{I_P}\mathtt{S_P}} \, \pi_{p}(\pi_{pe}(\mathbf{T}_{\mathtt{S}\mathtt{W}} \, \boldsymbol{\mu}_i))
\end{equation}

\paragraph{Elevation-Azimuth Image}
For elevation-azimuth images, the projected mean is obtained by:
\begin{equation}
 _\mathtt{I_E}\boldsymbol{\mu}_{i} = \mathbf{S}_{\mathtt{I_E}\mathtt{S_E}} \, \pi_{e}(\pi_{pe}(\mathbf{T}_{\mathtt{S}\mathtt{W}} \, \boldsymbol{\mu}_i))
\end{equation}

\subsubsection{Covariance Projection}

To propagate uncertainty from the world frame through the projection pipeline, we compute the projected covariance in image space for both polar and elevation-azimuth representations. 
The transformation matrices used in covariance projection exclude the translation components, denoted by $\hat{\mathbf{S}}_{\mathtt{I_P}\mathtt{S_P}}$ and $\hat{\mathbf{S}}_{\mathtt{I_E}\mathtt{S_E}}$, as translations do not affect the spread of uncertainty.

\paragraph{Polar Image}

The projected covariance for polar images is computed as:
\begin{equation}
_\mathtt{I_P}\mathbf{\Sigma}_i = \hat{\mathbf{S}}_{\mathtt{I_P}\mathtt{S_P}} \mathbf{J}_{\pi_{p}} \mathbf{J}_{\pi_{pe}} \mathbf{R}_{\mathtt{S}\mathtt{W}} \mathbf{\Sigma}_i \mathbf{R}_{\mathtt{S}\mathtt{W}}^{\mathtt{T}} \mathbf{J}_{\pi_{pe}}^{\mathtt{T}} \mathbf{J}_{\pi_{p}}^{\mathtt{T}} \hat{\mathbf{S}}_{\mathtt{I_P}\mathtt{S_P}}^{\mathtt{T}}
\end{equation}
where $\hat{\mathbf{S}}_{\mathtt{I_P}\mathtt{S_P}}$ is the $2 \times 2$ upper-left submatrix of $\mathbf{S}_{\mathtt{I_P}\mathtt{S_P}}$ that excludes the translation component, $\mathbf{J}_{\pi_{pe}}$ is the Jacobian of the Cartesian to polar-elevation conversion function $\pi_{pe}$ (mapping from $_\mathtt{S}\mathbf{p}_i$ to $_\mathtt{S_{PE}}\mathbf{p}_i$), $\mathbf{J}_{\pi_{p}}$ is the Jacobian of the polar projection function $\pi_p$ (mapping from $_\mathtt{S_{PE}}\mathbf{p}_i$ to $_\mathtt{S_P}\hat{\mathbf{p}}_i$), $\mathbf{R}_{\mathtt{S}\mathtt{W}}$ is the rotation from world to sonar frame, and $\mathbf{\Sigma}$ is the covariance in the world frame.

\paragraph{Elevation-Azimuth Image}

The projected covariance for elevation-azimuth images is computed as:
\begin{equation}
_\mathtt{I_E}\mathbf{\Sigma}_i = \hat{\mathbf{S}}_{\mathtt{I_E}\mathtt{S_E}} \mathbf{J}_{\pi_{e}} \mathbf{J}_{\pi_{pe}} \mathbf{R}_{\mathtt{S}\mathtt{W}} \mathbf{\Sigma}_i \mathbf{R}_{\mathtt{S}\mathtt{W}}^{\mathtt{T}} \mathbf{J}_{\pi_{pe}}^{\mathtt{T}} \mathbf{J}_{\pi_{e}}^{\mathtt{T}} \hat{\mathbf{S}}_{\mathtt{I_E}\mathtt{S_E}}^{\mathtt{T}}
\end{equation}
where $\hat{\mathbf{S}}_{\mathtt{I_E}\mathtt{S_E}}$ similarly represents the $2 \times 2$ upper-left submatrix of $\mathbf{S}_{\mathtt{I_E}\mathtt{S_E}}$ without the translation terms, $\mathbf{J}_{\pi_{pe}}$ is the Jacobian of the Cartesian to polar-elevation conversion function $\pi_{pe}$ (mapping from $_\mathtt{S}\mathbf{p}_i$ to $_\mathtt{S_{PE}}\mathbf{p}_i$), and $\mathbf{J}_{\pi_{e}}$ is the Jacobian of the elevation-azimuth projection function $\pi_e$ (mapping from $_\mathtt{S_{PE}}\mathbf{p}_i$ to $_\mathtt{S_E}\hat{\mathbf{p}}_i$).


\subsubsection{Two-Ways Splatting}
\begin{figure}
\centering
\includegraphics[width=0.92\linewidth]{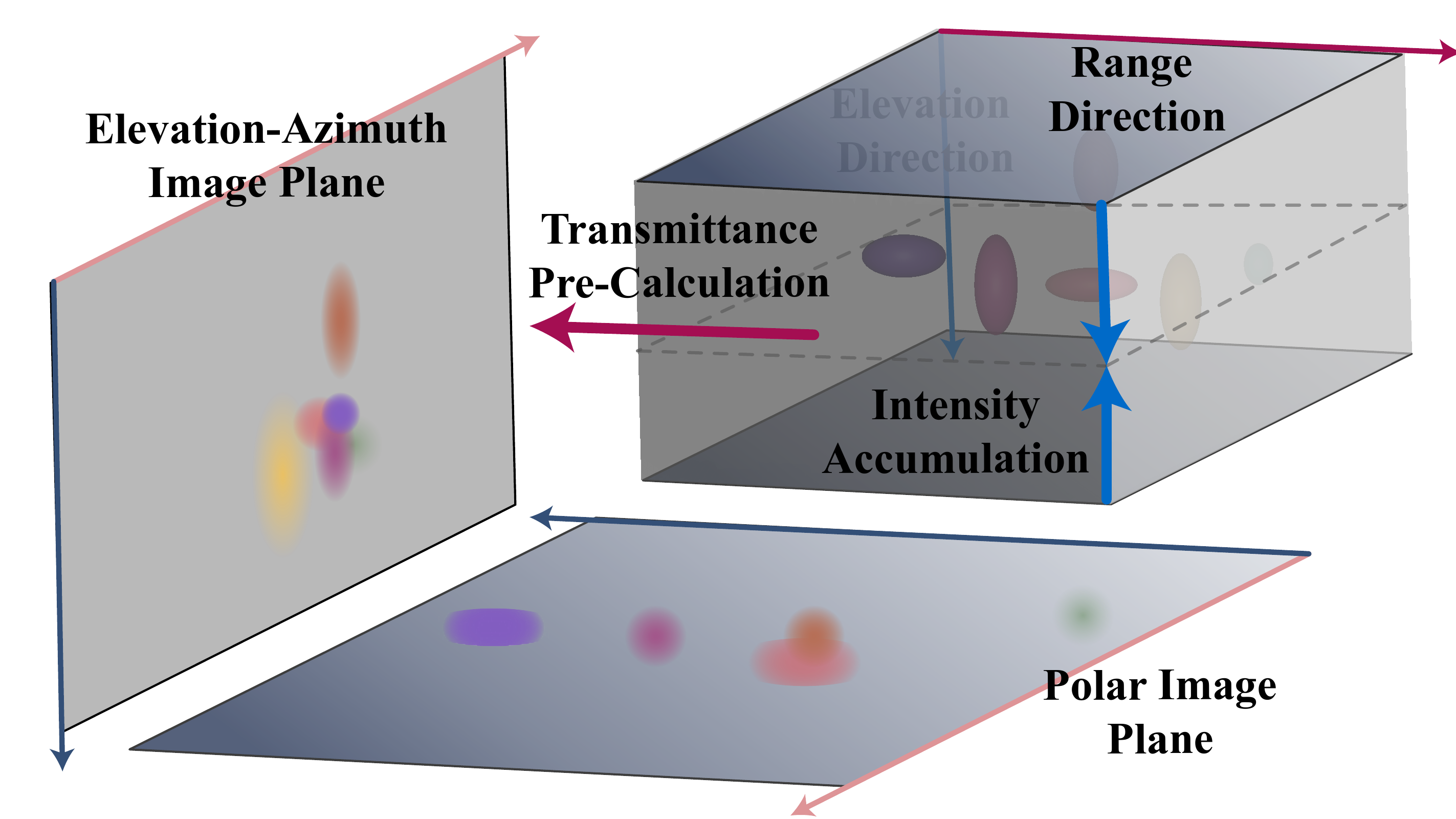}
\caption{Two-Ways Sonar Splat Visualization.}
\label{fig:sonar_splat}
\end{figure}

The critical differences between the proposed sonar Gaussian splatting and visual Gaussian splatting \cite{kerbl20233d} lie in the splatting and rendering process, which must account for the unique characteristics of sonar imaging. 

The original camera-based model performs splatting and alpha blending both along the depth (Z) direction, making transmittance calculation straightforward. However, in sonar imaging, the intensity values are accumulated along the elevation direction (corresponding to the polar image frame $\mathtt{I_P}$) while the transmittance calculation must be performed along the range direction (corresponding to the elevation-azimuth image frame $\mathtt{I_E}$). This dual-direction requirement introduces additional computational complexity during rendering.

To address this challenge, we propose Two-Ways Splatting, which incorporates the intensity accumulation and transmittance calculation simultaneously. The rendering process can be described as:
\begin{align}
  I &= \underbrace{\sum_{i\in\mathcal{N}}  I_i\alpha_i}_{\text{polar image frame $\mathtt{I_P}$}} \underbrace{\prod_{j\in\mathcal{M}}(1-\hat{\alpha}_j)}_{\text{elevation-azimuth image frame $\mathtt{I_E}$}}
  \label{equa:two_ways_splatting}
\end{align}
where $\alpha_i$ is the opacity of the $i$-th Gaussian evaluated on the 2D polar image frame $\mathtt{I_P}$, and $\mathcal{N}$ is the set of Gaussians projected onto the current pixel along the elevation arc. The term $\prod_{j \in \mathcal{M}}(1-\hat{\alpha}_j)$ represents the transmittance, which is pre-computed on the elevation-azimuth image frame $\mathtt{I_E}$, where $\mathcal{M}$ denotes the set of Gaussians located before the current Gaussian along the range direction, and $\hat{\alpha}_j$ is the opacity of the $j$-th Gaussian evaluated on the elevation-azimuth image frame.

In the proposed Two-Ways Splatting, we divide the splatting process into two steps:

First, we splat the Gaussians onto the frame $\mathtt{I_E}$ to pre-compute the transmittance along the range direction at each Gaussian's 2D mean position $_\mathtt{I_E}\boldsymbol{\mu}_i$. This procedure computes only one transmittance value per Gaussian, which is subsequently used during alpha blending. Although this approach approximates the transmittance for all pixels within a Gaussian's footprint, it significantly reduces computational cost compared to repeatedly calculating transmittance for each pixel during polar image splatting.

Second, after obtaining the transmittance, we perform alpha blending in the frame $\mathtt{I_P}$. This process follows the standard Gaussian Splatting pipeline to accumulate intensity along the elevation direction while incorporating the pre-computed transmittance values as in \eqref{equa:two_ways_splatting}.

\subsection{Sonar Noise Modeling with Gaussian Mixture Models}

Sonar imaging is inherently affected by complex noise patterns that arise from the interaction between acoustic beams and the underwater environment. To model this noise, we employ a Gaussian Mixture Model (GMM) that captures the characteristics of sonar beam patterns in both azimuth and range dimensions as shown in Figure \ref{fig:gmm}. Our approach learns image-specific noise distributions that can be applied during rendering to synthesize realistic sonar imagery.

\subsubsection{GMM Formulation}

We model the noise distribution for each sonar image using two independent GMMs: one for the azimuth (cross-range) direction and one for the range (along-track) direction. For an image with index $n \in \{1,\ldots,N\}$, the complete noise model is defined by:

\begin{figure}
\centering
\includegraphics[width=0.92\linewidth]{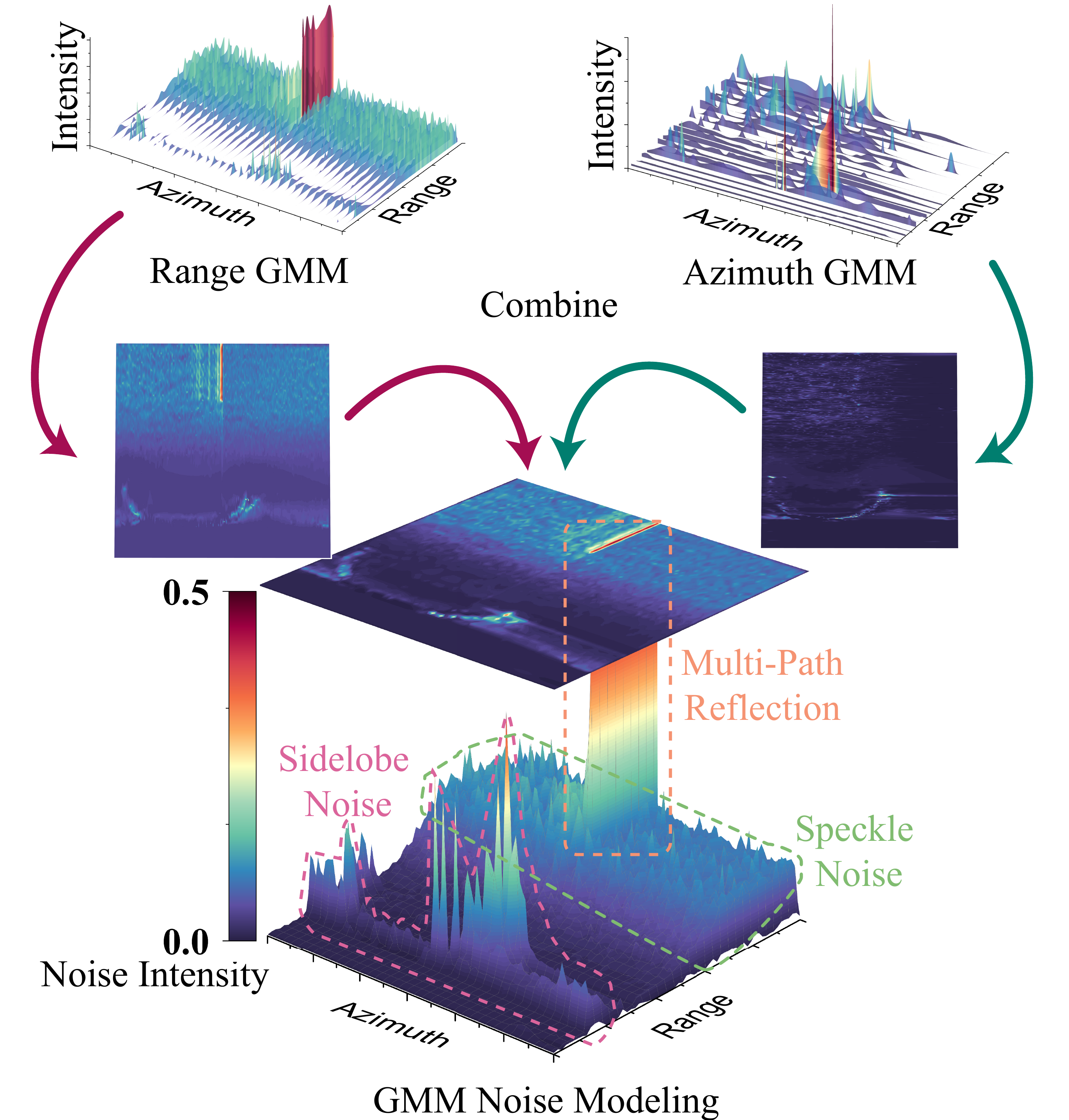}
\caption{Proposed GMM Noise Model for Sonar Imaging.}
\label{fig:gmm}
\end{figure}

\paragraph{Azimuth GMM}
The azimuth noise distribution is modeled per range bin using $K$ Gaussian components:
\begin{equation}
p(\theta | h, n) = \sum_{k=1}^{K} \pi_{k,h}^{(n)} \mathcal{N}(\theta; \mu_{k,h}^{(n)}, (\sigma_{k,h}^{(n)})^2)
\end{equation}
where:
\begin{itemize}
    \item $\theta \in [-\text{FOV}/2, \text{FOV}/2]$ is the azimuth angle
    \item $h \in \{1,\ldots,H\}$ is the range bin index ($H$ in total)
    \item $\mu_{k,h}^{(n)} \in \mathbb{R}$ and $\sigma_{k,h}^{(n)} \in \mathbb{R}^+$ are the mean and the standard deviation for component $k$ at row $h$
    \item $\pi_{k,h}^{(n)} \in [0,1]$ is the mixing coefficient with $\sum_{k=1}^{K} \pi_{k,h}^{(n)} = 1$
\end{itemize}


Additionally, a gain parameter $g_h^{(n)} \in \mathbb{R}^+$ per range bin modulates the overall noise intensity.

\paragraph{Range GMM}
Similarly, the range noise distribution is modeled per azimuth bin using $K$ Gaussian components:
\begin{equation}
p(r | w, n) = \sum_{k=1}^{K} \tilde{\pi}_{k,w}^{(n)} \mathcal{N}(r; \tilde{\mu}_{k,w}^{(n)}, (\tilde{\sigma}_{k,w}^{(n)})^2)
\end{equation}
where:
\begin{itemize}
    \item $r \in [0, R_{\max}]$ is the range distance
    \item $w \in \{1,\ldots,W\}$ is the azimuth bin index ($W$ in total)
    \item $\tilde{\mu}_{k,w}^{(n)} \in \mathbb{R}$ and $\tilde{\sigma}_{k,w}^{(n)} \in \mathbb{R}^+$ are the mean range and the standard deviation for component $k$ at column $w$
    \item $\tilde{\pi}_{k,w}^{(n)} \in [0,1]$ is the mixing coefficient with $\sum_{k=1}^{K} \tilde{\pi}_{k,w}^{(n)} = 1$
\end{itemize}
Similarly, it is associated with a gain parameter $\tilde{g}_w^{(n)}$ learned per azimuth bin.

\subsubsection{Noise Generation}

When rendering a pixel $(n,h,w)$ at a particular position on an image,  we sample noise values from both azimuth and range GMMs to generate the final noise contribution.

\paragraph{Azimuth Noise}
\begin{equation}
N_\theta(h,w) = g_h^{(n)} \sum_{k=1}^{K} \pi_{k,h}^{(n)} \exp\left(-\frac{(\theta_w - \mu_{k,h}^{(n)})^2}{2(\sigma_{k,h}^{(n)})^2}\right)
\end{equation}

\paragraph{Range Noise}
\begin{equation}
N_r(h,w) = \tilde{g}_w^{(n)} \sum_{k=1}^{K} \tilde{\pi}_{k,w}^{(n)} \exp\left(-\frac{(r_h - \tilde{\mu}_{k,w}^{(n)})^2}{2(\tilde{\sigma}_{k,w}^{(n)})^2}\right)
\end{equation}

\paragraph{Combined Rendering}
The final rendered image combines the clean splatted output $I_{\text{clean}}$ with both noise components:
\begin{equation}
I_{\text{final}}(h,w) = I_{\text{clean}}(h,w) + N_\theta(h,w) + N_r(h,w)
\end{equation}

\begin{figure*}
\centering
\includegraphics[width=2\columnwidth]{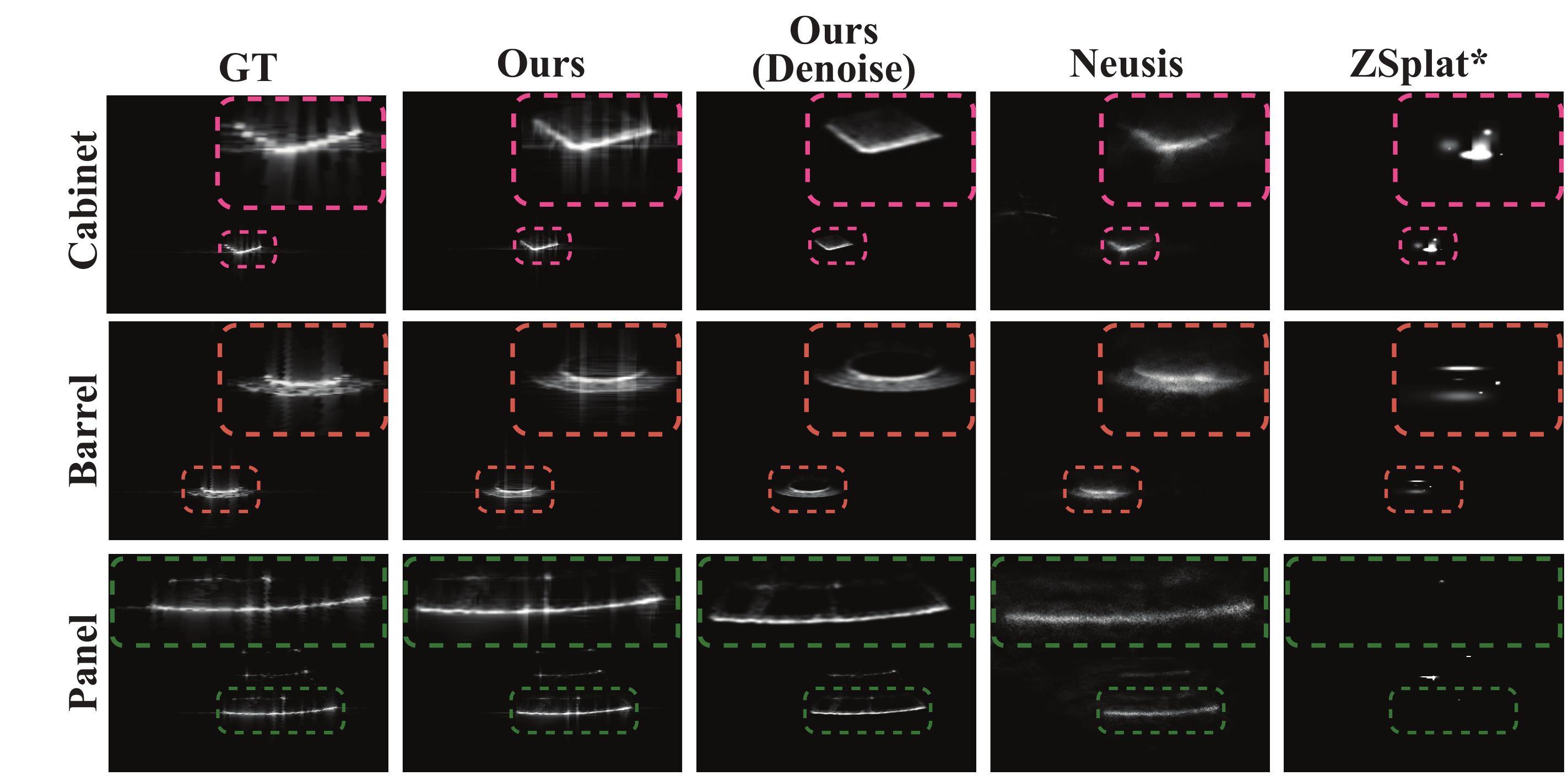}
\caption{Qualitative Comparison of Rendered Sonar Images on the Simulation Dataset.}
\label{fig:rendering_comparison}
\end{figure*}

\subsection{Pipeline Overview}
Figure \ref{fig:pipeline} illustrates the overall pipeline of the proposed NAS-GS method. Our training can be divided into two stages. The first stage focuses on initially optimizing the 3D GS for the scene geometry. The second stage incrementally incorporates the proposed GMM-based noise model to refine the Gaussians, while learning the GMM noise distributions. 

In the first stage, 
we first initialize GS from polar images where pixel intensity is greater than a threshold. Then we project these GS into the sonar image space using the Sonar Geometry transformations. The rendering is performed with the proposed Two-Ways Splatting technique that first calculates transmittance along range directions for each GS efficiently before accumulating pixel intensities along elevation direction. Different from the original 3D GS \cite{kerbl20233d} which uses single-view rendering loss to optimize the GS, we use a multi-view loss. This is because the scene geometry is consistently represented across views in sonar imaging, while complex noise randomly appears depending on the viewpoint. Therefore, using multi-view loss can help the Gaussians focus on fitting the underlying geometry rather than noises in individual views.

In the second stage, the GMM-based noise model is introduced to capture sonar noises.  Each model is parameterized by learnable means, variances, and mixing coefficients for GMM components. During training, we jointly optimize the 3D GS parameters and the GMM parameters using a combined loss function that includes a reconstruction loss between rendered and reference images. This allows the GS to fit the true scene geometry while the GMM captures the complex noise characteristics inherent in sonar imaging.

\section{Experimental Results}


\subsection{Environment Setup}

\subsubsection{Datasets}
Both simulated and real-world offshore underwater sonar datasets are used for evaluation. 

\paragraph{Simulation Dataset}
The simulation dataset is generated using the ray-based multi-beam sonar simulator \cite{choi2021physics} since it can produce high-fidelity sonar images. 
Three different scenes (Cabinet, Barrel, Panel) are created in the simulator, each of which is scanned using a Blueview p900 sonar model with 90 degree horizontal field of view and 20 degree vertical field of view. The sonar is equipped on a robot platform traveling around four sides of the objects and captures over 100 sonar images with a resolution of 512x399 pixels.
The sonar pose data is also obtained from the simulator. 

\paragraph{Offshore Dataset}
The offshore dataset was collected using an underwater robot equipped with a Tritech Gemini 1200ik imaging sonar at an offshore wind turbine foundation with a height of approximately 50 meters. The robot is also equipped with a stereo camera, an IMU, and a DVL. A long sequence with a over 20 minutes return trajectory from the surface to the seabed is used to cover various types of real noise (side-lobe, speckle, and multi-path noise) present in the sonar images for evaluation in challenging offshore large-scale scenarios. The same set of 6-DoF poses estimated through Aqua-SLAM \cite{xu2025aqua} is used for training. 

\subsubsection{Competing Methods and Evaluation Metrics}

We compare our method against two baseline approaches. Neusis \cite{qadri2023neural} is a neural implicit surface reconstruction method designed for sonar data that uses signed distance functions and volume rendering. ZSplat$^{\star}$ \cite{qu2024z} is an adaptation of the camera-sonar 3D Gaussian Splatting method whose camera model is disabled for sonar-only evaluation. 

\subsubsection{Gaussian Initialization}
For the simulation dataset, we use simulated lidar scans to generate initial Gaussians for NAS-GS and ZSplat$^{\star}$ training since ZSplat requires initialization from point clouds. However, our method supports initialization directly from sonar images. Therefore, in the offshore dataset, we directly initialize Gaussians using sonar images with pixel intensity above a certain threshold to demonstrate the practicality of our approach in real-world scenarios where prior point cloud may not be available.

\subsubsection{Evaluation Metrics}
For novel view synthesis, rendering quality is evaluated using the standard metrics introduced in \cite{kerbl20233d}: Peak Signal-to-Noise Ratio (PSNR), Structural Similarity Index (SSIM) and Learned Perceptual Image Patch Similarity (LPIPS).
For 3D reconstruction, Chamfer Distance (CD) and Hausdorff Distance (HD) \cite{qadri2023neural} are used to assess 3D geometric accuracy. 

\subsubsection{Hardware}
All experiments are conducted on a desktop PC equipped with a NVIDIA RTX 4070 GPU (12GB), AMD Ryzen 7 7800X3D CPU and 32GB RAM.

\begin{table}
  \centering
  \captionsetup{font=small}
  \caption{Evaluation on Novel View Synthesis. }
  \label{table:novel_view_synthesis}
  \footnotesize
  \setlength{\tabcolsep}{1pt}
  \begin{tabular*}{\columnwidth}{@{\extracolsep{\fill}}l|ccc|ccc|ccc@{}}
    \toprule
    & \multicolumn{3}{c|}{\textbf{Ours}} 
    & \multicolumn{3}{c|}{\textbf{Neusis}} 
    & \multicolumn{3}{c}{\textbf{ZSplat}$^{\star}$} \\
    \midrule
    \textbf{Scene} & \fontsize{7}{4}\selectfont PSNR\!$\uparrow$ & \fontsize{7}{4}\selectfont SSIM\!$\uparrow$ & \fontsize{7}{4}\selectfont LPIPS\!$\downarrow$
                  & \fontsize{7}{4}\selectfont PSNR\!$\uparrow$ & \fontsize{7}{4}\selectfont SSIM\!$\uparrow$ & \fontsize{7}{4}\selectfont LPIPS\!$\downarrow$
                  & \fontsize{7}{4}\selectfont PSNR\!$\uparrow$ & \fontsize{7}{4}\selectfont SSIM\!$\uparrow$ & \fontsize{7}{4}\selectfont LPIPS\!$\downarrow$ \\
    \midrule
    \textbf{Cabinet} & \best{38.95} & \best{0.99} & \best{0.03} & \secondbest{32.78} & 0.92 & 0.09 & 29.73 & \secondbest{0.97} & \secondbest{0.05} \\
    \textbf{Barrel} & \best{37.95} & \best{0.98} & \best{0.04} & \secondbest{30.87} & 0.89 & 0.10 & 28.73 & \secondbest{0.97} & \secondbest{0.05} \\
    \textbf{Panel} & \best{37.42} & \best{0.98} & \best{0.04} & \secondbest{29.79} & 0.85 & 0.17 & 27.04 & \secondbest{0.96} & \secondbest{0.08} \\
    \bottomrule
  \end{tabular*}
  
  \small
  \vspace{2pt}
  \textit{Note:} \textbf{Best} results are bolded, \underline{second-best} results are underlined.
\end{table}

\subsection{Evaluation using Simulation Data}
\subsubsection{Sonar Image Rendering and View Synthesis}
The quantitative results of the sonar novel view synthesis performance are summarized in Table \ref{table:novel_view_synthesis}, and the qualitative comparisons are shown in Figure \ref{fig:rendering_comparison}.
Our method achieves superior performance across all metrics compared to the baselines. We consistently obtain PSNR values above 37 dB, SSIM scores near 0.98, and LPIPS values below 0.05, demonstrating both high pixel-wise accuracy and perceptual quality. The "Ours (Denoise)" variant, which removes the learned GMM noise during rendering, shows that our GMM-based noise model effectively captures the sonar-specific artifacts and helps the 3D Gaussian representation focus on the real geometry.
Neusis does not capture fine details due to its lack of noise handling, resulting in PSNR values around 30-33 dB and noticeably lower SSIM scores. ZSplat$^{\star}$, while achieving better structural similarity than Neusis, shows sparse and blurred rendering results with PSNR values around 27-30 dB compared to our method.
The qualitative results in Figure \ref{fig:rendering_comparison} clearly demonstrate the effectiveness of the proposed approach. It preserves sharp edges and fine structures while accurately reproducing the characteristic sonar noise patterns. The denoised version reveals the underlying geometric structure rendered by the 3D GS beneath the learned GMM noise model.

\subsubsection{Sonar based 3D Reconstruction}
To convert the 3D Gaussian representation to a mesh, we first extract a dense point cloud by sampling points from each Gaussian distribution within three standard deviations of its mean, retaining only points where the evaluated density exceeds a predefined threshold. We then apply the marching cubes algorithm \cite{lorensen1998marching} to reconstruct a surface mesh from this point cloud. 

The quantitative results comparing the reconstructed meshes against ground truth models are presented in Table \ref{table:3d_reconstruction}.
Our method demonstrates accurate geometric reconstruction across all scenes, achieving Chamfer Distances on the order of 0.005-0.060 and Hausdorff Distances below 0.5, representing an order of magnitude improvement over the baseline methods in most cases.
Neusis and ZSplat$^{\star}$ produce reconstructions with substantial geometric errors. 

\begin{table}
  \centering
  \captionsetup{font=small}
  \caption{Evaluation on 3D Reconstruction}
  \label{table:3d_reconstruction}
  \footnotesize
  \setlength{\tabcolsep}{6pt}
  \begin{tabular*}{\columnwidth}{@{\extracolsep{\fill}}c|cc|cc|cc@{}}
    \toprule
    & \multicolumn{2}{c|}{\textbf{Ours}} 
    & \multicolumn{2}{c|}{\textbf{Neusis}} 
    & \multicolumn{2}{c}{\textbf{ZSplat$^{\star}$}} \\ 
    \midrule
    \textbf{Scene} & CD$\downarrow$ & HD$\downarrow$
                  & CD$\downarrow$ & HD$\downarrow$
                  & CD$\downarrow$ & HD$\downarrow$ \\
    \midrule
    Cabinet & \best{0.006} & \best{0.054} & \secondbest{0.277} & 1.333 & 0.298 & \secondbest{0.871} \\
    Barrel & \best{0.007} & \best{0.012} & 2.096 & 6.676 & \secondbest{0.020} & \secondbest{0.320} \\
    Panel & \best{0.060} & \best{0.464} & \secondbest{0.313} & 2.633 & 0.342 & \secondbest{1.249} \\
    \bottomrule
  \end{tabular*}
  
  \small
  \vspace{2pt}
  \textit{Note:} CD = Chamfer Distance, HD = Hausdorff Distance. \textbf{Best} results are bolded, \secondbest{second-best} results are underlined.
\end{table}

\begin{figure}
\centering
\setlength{\tabcolsep}{1pt} 

\begin{adjustbox}{max width=\linewidth}
\begin{tabular}{>{\centering\arraybackslash}m{0.11\linewidth} c c c c}
\textbf{} & \textbf{GT} & \textbf{Ours} & \textbf{Neusis} & \textbf{ZSplat$^{\star}$} \\

\rotatebox{90}{\textbf{\makecell[c]{Cabinet}}} &
\vci[width=0.17\linewidth]{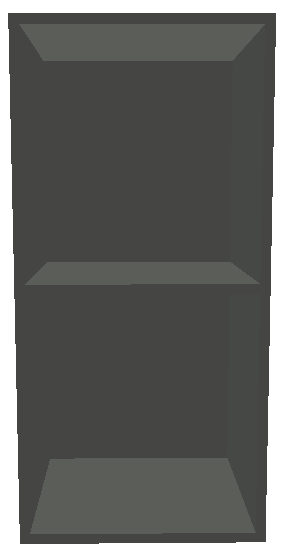} &
\vci[width=0.17\linewidth]{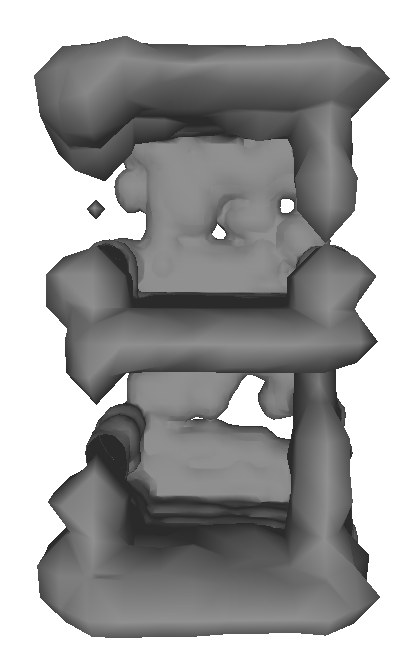} &
\vci[width=0.17\linewidth]{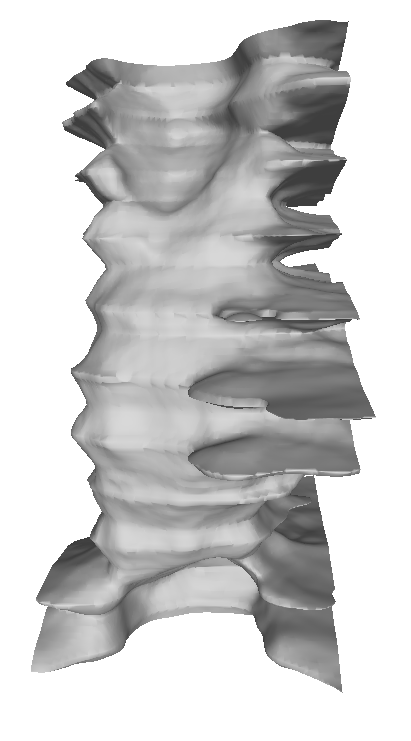} &
\vci[width=0.17\linewidth]{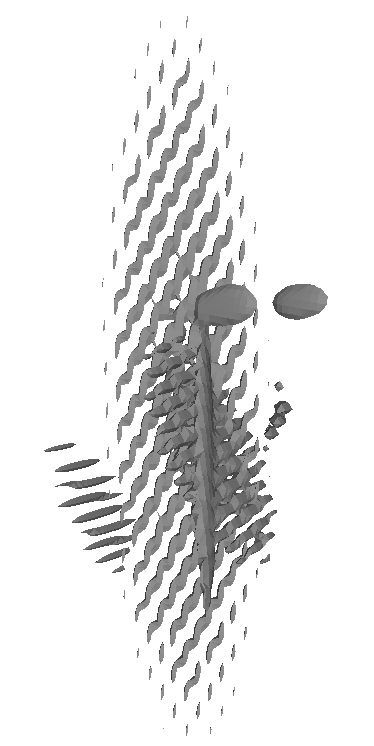} \\

\rotatebox{90}{\textbf{\makecell[c]{Barrel}}} &
\vci[width=0.17\linewidth]{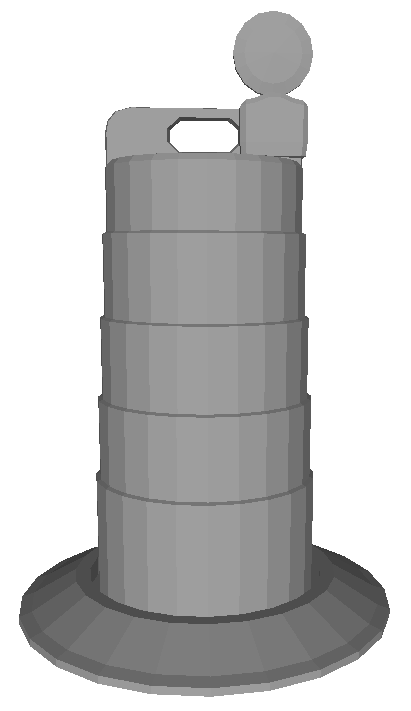} &
\vci[width=0.17\linewidth]{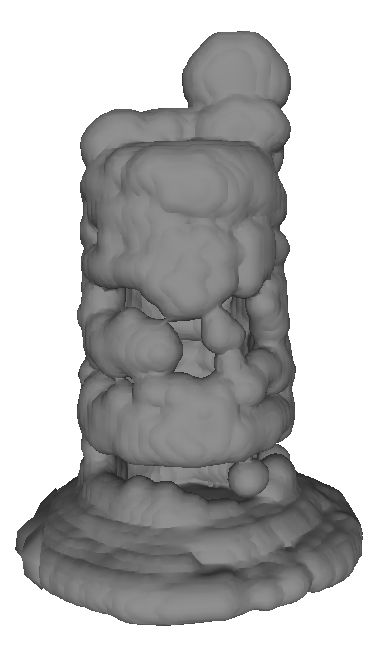} &
\vci[width=0.17\linewidth]{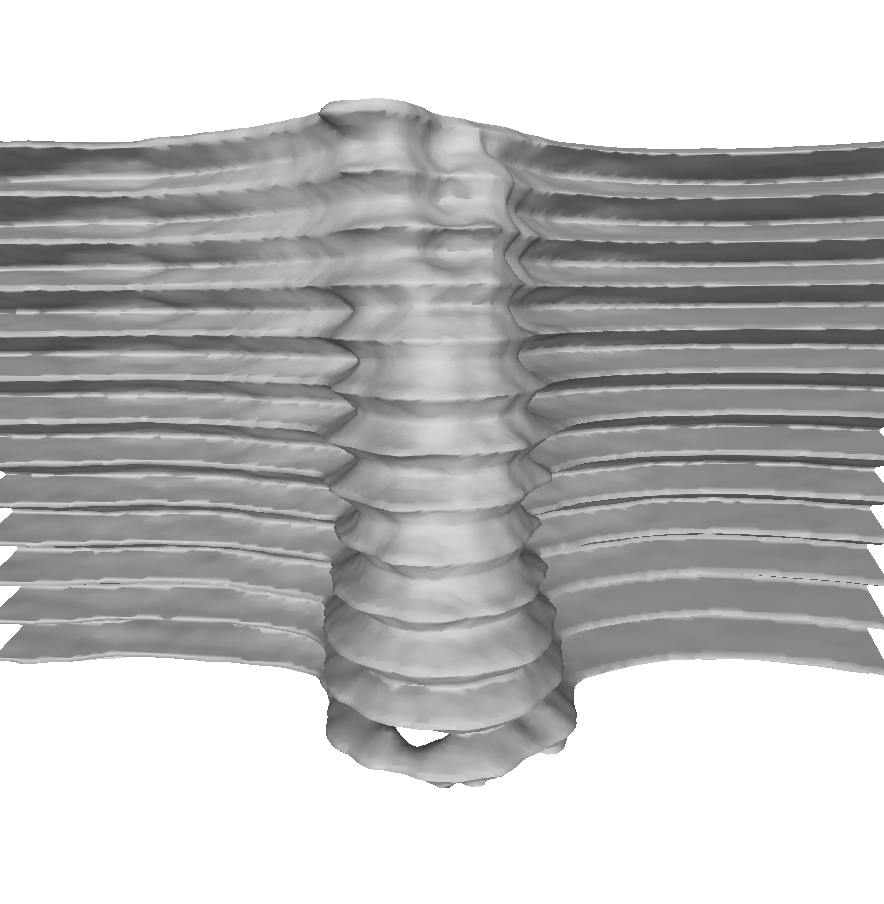} &
\vci[width=0.17\linewidth]{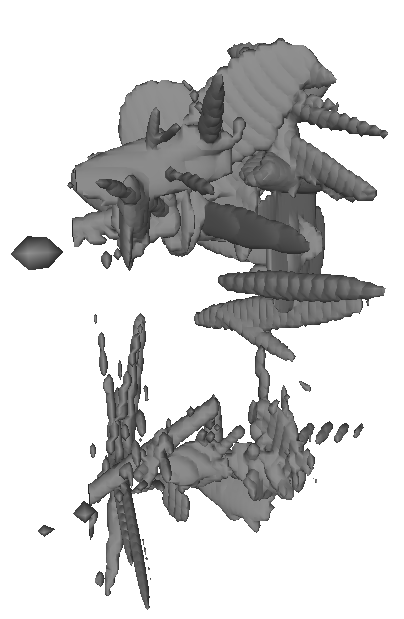} \\

\rotatebox{90}{\textbf{\makecell[c]{Panel}}} &
\vci[width=0.17\linewidth]{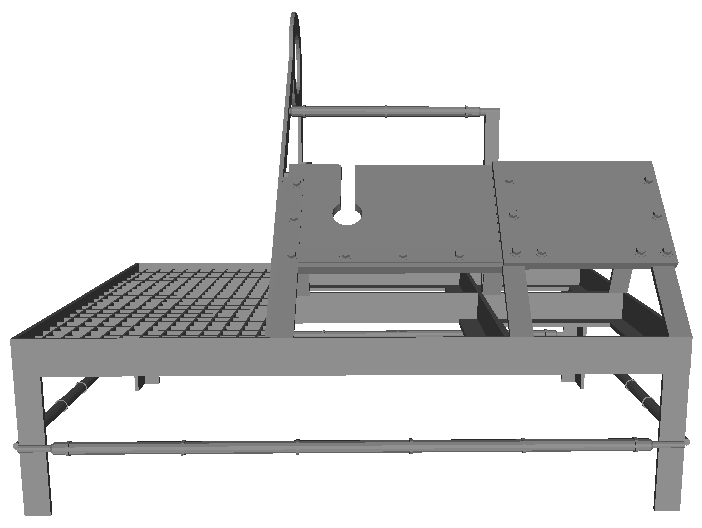} &
\vci[width=0.17\linewidth]{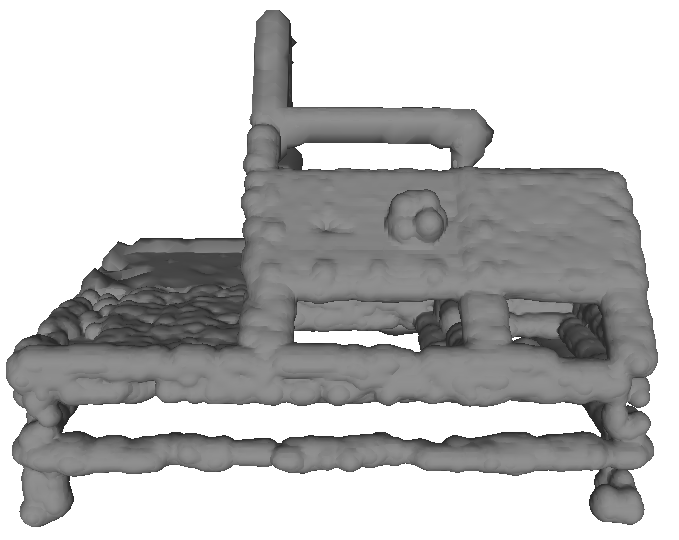} & 
\vci[width=0.17\linewidth]{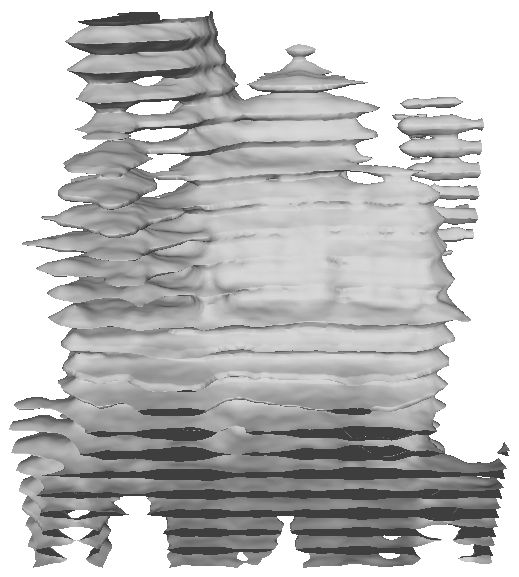} &
\vci[width=0.17\linewidth]{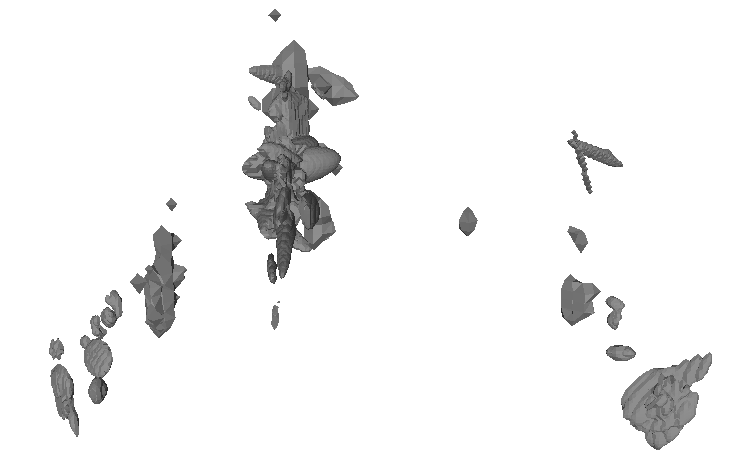} \\

\end{tabular}
\end{adjustbox}

\caption{Qualitative Comparison of Sonar 3D reconstruction.}
\label{fig:mesh_comparison}
\end{figure}

The qualitative mesh comparisons in Figure \ref{fig:mesh_comparison} further illustrate these differences. Our reconstructed meshes exhibit surface structures with reasonable geometric details that match the ground truth models. In contrast, Neusis meshes show structures with wave-shaped distortions on the edges, overfitting to the noise, while ZSplat$^{\star}$ meshes contain visible holes and irregular surfaces due to the lack of handling sonar-specific imaging characteristics.

\subsection{Evaluation using Real Offshore Data}

We further validate the proposed method using real-world offshore data for sonar based novel-view synthesis and large-scale reconstruction. The sequence spans over 20 minutes and covers a large-scale trajectory around an foundation of an wind turbine located North Sea. This challenging scenario features various types of complex noise patterns, including side-lobe artifacts, speckle noise, and multi-path reflections that are characteristic of real offshore environments. 

We compared our sonar 3D reconstruction results against the vision-based reconstruction using Block Matching(BM) stereo matching. Notably, we only initialize GS from sonar images where intensity is greater than a threshold, without using any prior point cloud. And poses from Aqua-SLAM \cite{xu2025aqua} are used for both methods. 








\begin{figure}
\centering
\includegraphics[width=1\columnwidth]{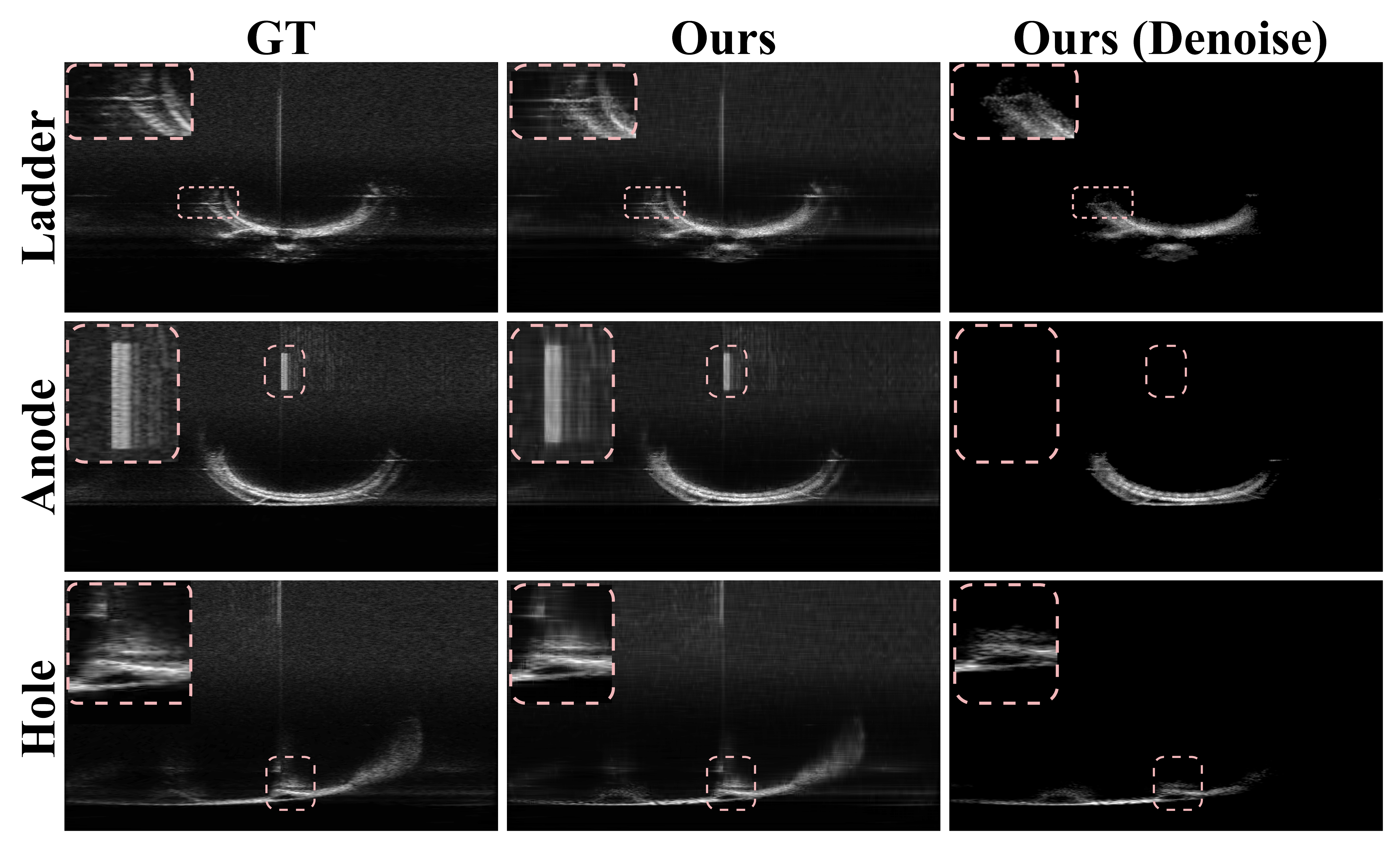}
\caption{Offshore novel view synthesis results.}
\label{fig:offshore_rendering}
\end{figure}

Figure \ref{fig:offshore_rendering} shows representative frames from three distinct regions: a ladder structure on the foundation, an anode, and a towing hole structure. The proposed method can render realistic sonar images that visually match the ground truth observations. Meanwhile, the denoised version reveals the clean geometric structure, demonstrating that our GMM-based noise model effectively separates the true geometry from sonar-specific artifacts.

\begin{figure}
\centering
\includegraphics[width=\linewidth]{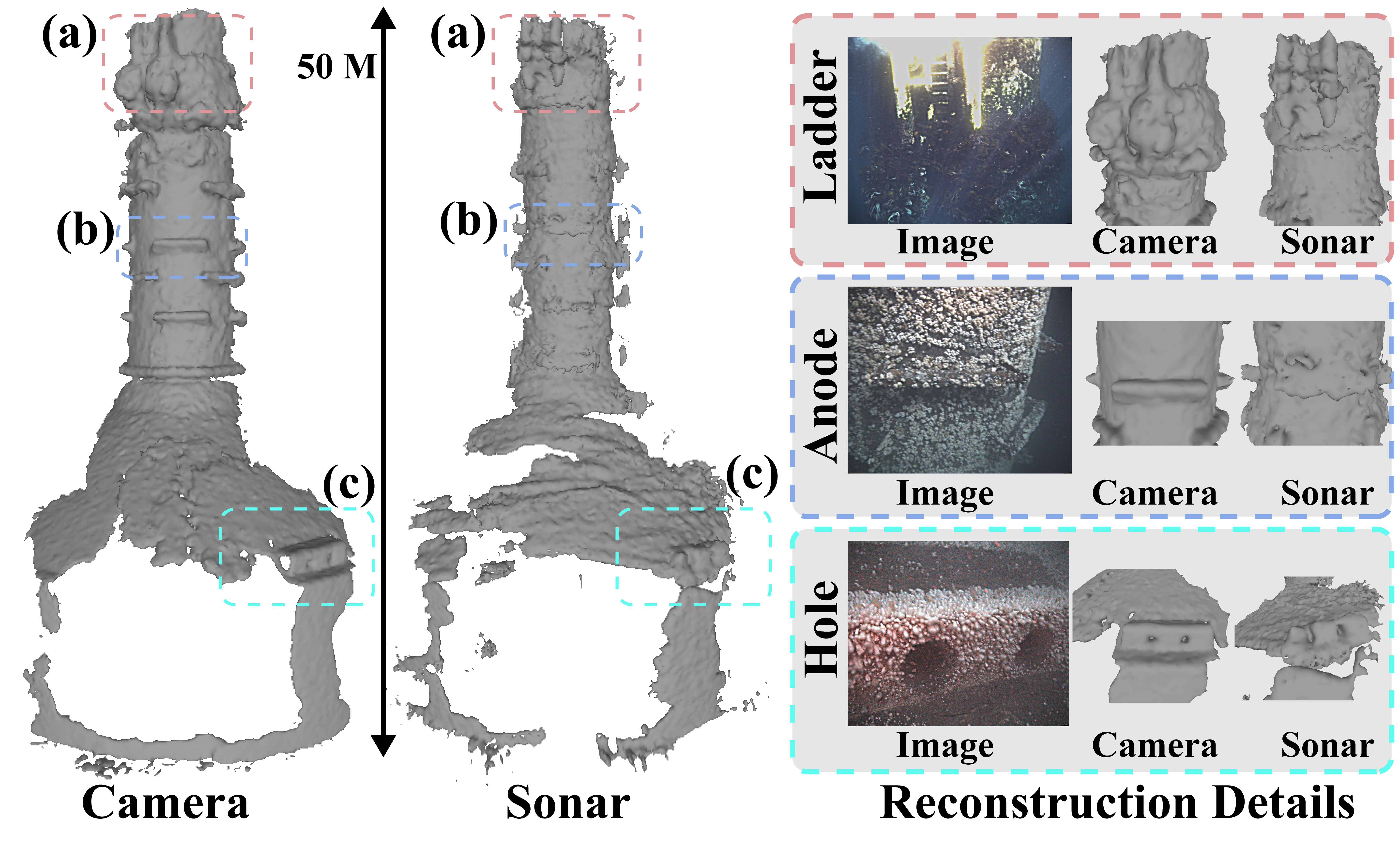}
\caption{Sonar vs Camera Reconstruction on Offshore Dataset. Camera images are not used for sonar reconstruction.}
\label{fig:offshore}
\end{figure}

Figure \ref{fig:offshore} presents the sonar-only reconstructed 3D model of the structure, compared with vision-based reconstruction \cite{xu2025aqua}. Despite the challenging conditions and large scale of the scene, our reconstruction captures the complex geometric features of the underwater structure, including the ladder, anodes, and various structural details. 
Notably, the reconstruction of the algae-covered ladder demonstrates superior quality compared to vision-based results. This is because the vision-based method is compromised by the dynamic movement of the algae, whereas the acoustic noise from the algae is minimal and effectively modeled by our GMM, resulting in a cleaner reconstruction from sonar data.

\subsection{Computation}

We evaluate the computational efficiency of our method by measuring the rendering speed in FPS on the simulation dataset. The rendering speed is highly dependent on the number of 3D Gaussians used to represent the scene. In our experiments, we load the trained models with approximately 10000 Gaussians which already provide high-quality rendering and reconstruction results for our method. ZSplat loads the same ply file to ensure the same number of Gaussians are used. Table \ref{tab:fps_comparison} presents the rendering performance comparison across all three methods.

Our method achieves real-time rendering ($>$700 FPS), approx. 2$\times$ faster than ZSplat$^{\star}$ and 8,000$\times$ faster than Neusis. Ours$^{\star}$ (pure GS without GMM) reaches $\sim$1000 FPS. This efficiency stems from our Two-Ways Splatting, which pre-computes transmittance in the elevation-azimuth frame before fast alpha blending in the polar frame. In contrast, ZSplat$^{\star}$ computes transmittance repeatedly for each pixel during rendering, while Neusis requires computationally expensive ray marching through implicit surface representations. The superior rendering efficiency of our method makes it particularly suitable for large-scale offshore environments.

\begin{table}
  \centering
  \caption{Rendering Performance Comparison}
  \label{tab:fps_comparison}
  \footnotesize
  \setlength{\tabcolsep}{10pt}
  \renewcommand{\arraystretch}{1.2}
  \begin{tabular*}{\columnwidth}{@{\extracolsep{\fill}}c|c|c|c|c@{}}
    \toprule
    \textbf{Scene} & \textbf{Ours} & \textbf{Ours}$^{\star}$ & \textbf{Neusis} & \textbf{ZSplat$^{\star}$} \\
     & \textbf{FPS}$\uparrow$ & \textbf{FPS}$\uparrow$ & \textbf{FPS}$\uparrow$ & \textbf{FPS}$\uparrow$ \\
    \midrule
    Cabinet & \secondbest{735.98} & \best{962.01} & 0.09 & 411.62 \\
    Barrel & \secondbest{832.36} & \best{1032.04} & 0.08 & 419.45 \\
    Panel & \secondbest{803.12} & \best{1274.23} & 0.08 & 417.66 \\
    \bottomrule
  \end{tabular*}
  
  \small
  \vspace{2pt}
  \textit{Note:} \textbf{Ours} stands for full model, \textbf{Ours$^{\star}$} stands for w/o GMM. \textbf{Best} results are bolded, \secondbest{second-best} results are underlined.
\end{table}

\subsection{Ablation Study}
To further analyze the contributions of our proposed components, we conduct an ablation study on the simulation dataset. We evaluate the impact of the GMM-based noise modeling on both novel view synthesis and 3D reconstruction performance. 
\subsubsection{Novel View Synthesis Ablation}

\begin{figure}
\centering
\includegraphics[width=\linewidth]{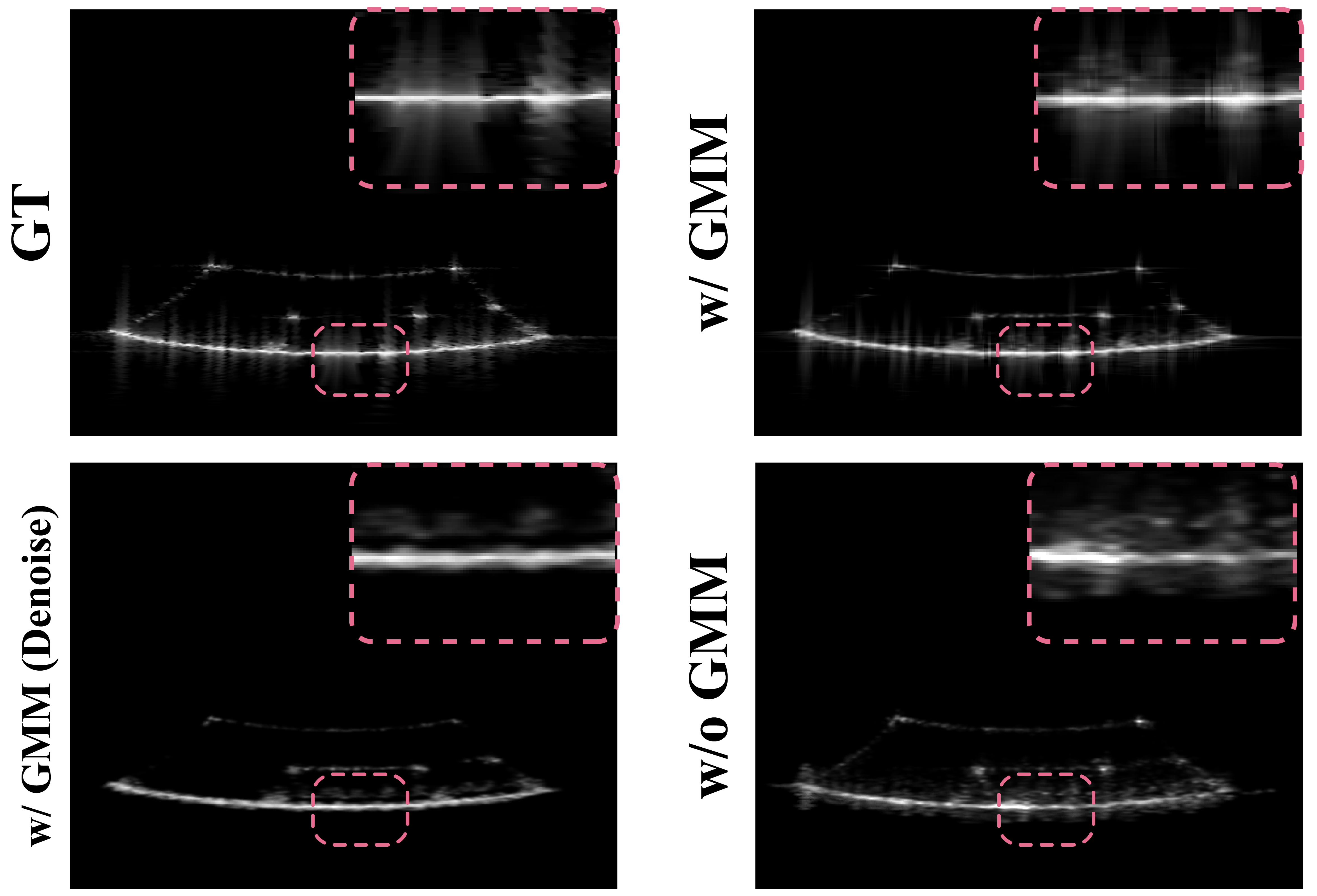}
\caption{Ablation study on GMM-based noise modeling for novel view synthesis.}
\label{fig:ablation_novel_view_synthesis}
\end{figure}

As shown in Figure \ref{fig:ablation_novel_view_synthesis}, we compare the full NAS-GS method with a variant that excludes the GMM-based noise modeling. The results clearly demonstrate that the GMM noise model significantly enhances rendering quality. The rendered image with GMM closely matches the ground truth, and the denoised rendering (removing GMM noise to isolate pure GS output) confirms that the GMM effectively captures sonar-specific noise patterns while allowing 3D Gaussians to focus on representing the underlying scene structure. Without the GMM noise model, rendering quality degrades substantially. Since sonar noise varies randomly across viewpoints, the 3D Gaussians alone cannot consistently model noise across all views, resulting in blurred and distorted renderings. 

\subsubsection{3D Reconstruction Ablation}

\begin{figure}
\centering
\includegraphics[width=\linewidth]{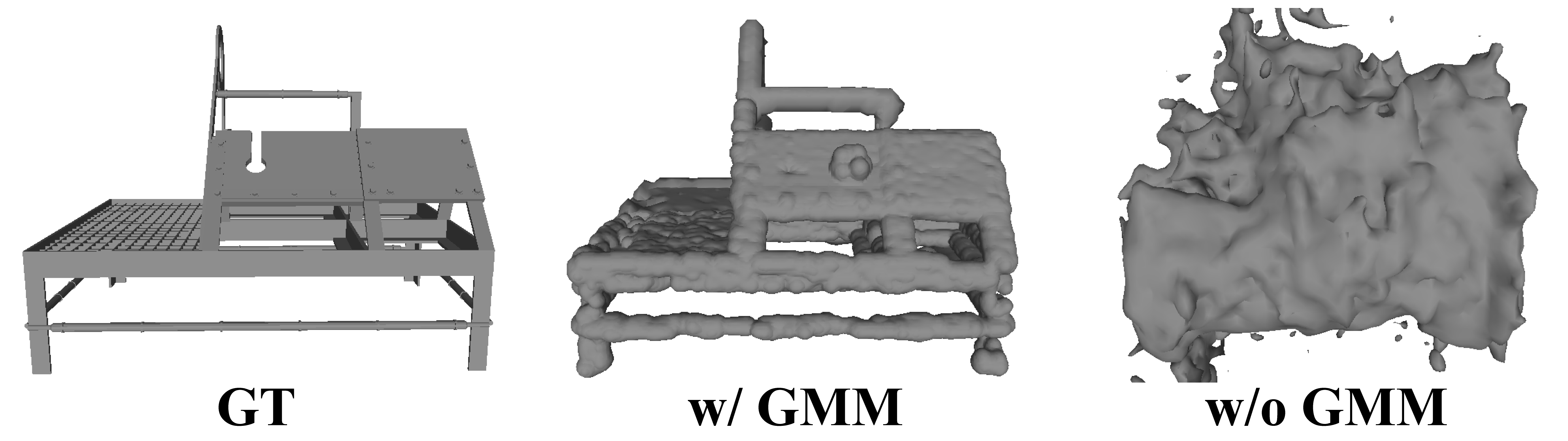}
\caption{Ablation study on GMM-based noise modeling on 3D reconstruction.}
\label{fig:ablation_3d_reconstruction}
\end{figure}

As shown in Figure \ref{fig:ablation_3d_reconstruction}, we evaluate the impact of GMM-based noise modeling on 3D reconstruction accuracy. Without the GMM noise model, reconstruction quality deteriorates, exhibiting distorted surfaces as the 3D Gaussians overfit to noise. This highlights the importance of accurately modeling sonar noise, enabling 3D Gaussians to focus on capturing true geometric features rather than overfitting to noise artifacts. The results indicate that GMM noise modeling contributes significantly to improved reconstruction performance.

\section{Conclusions}

This paper introduces NAS-GS, a noise-aware sonar Gaussian splatting framework that addresses the unique challenges of sonar reconstruction and novel view synthesis. The proposed two-ways Splatting technique efficiently handles sonar's polar imaging geometry, while the GMM-based noise modeling effectively captures complex sonar-specific noise patterns and allow the 3D Gaussians to focus on representing the underlying scene structure. Extensive experiments on both simulated and real-world offshore datasets demonstrate that NAS-GS outperforms state-of-the-art methods in novel view synthesis and 3D reconstruction tasks, achieving high-fidelity results with real-time rendering speeds. Future work will explore extending the framework to sonar SLAM and dynamic scenes and integrating temporal coherence for improved reconstruction of time-varying underwater environments.


\bibliographystyle{IEEEtran}
\bibliography{./ref}

\end{document}